\documentclass[12pt, final]{l4dc2022}

\usepackage{graphicx}
\usepackage{times}
\usepackage{amsmath}
\usepackage{amssymb}
\usepackage{amsfonts}
\usepackage[utf8]{inputenc} 
\usepackage{xspace}

\usepackage{rgap_notation}
\usepackage{rgap_header}

\newcommand*{\field}[1]{\mathbb{\MakeUppercase{#1}}}		

\newcommand*{\sequence}[1]{\mathrm{\MakeUppercase{#1}}}

\newcommand*{\R}{\field{R}} 

\renewcommand{\vec}[1]{{\boldsymbol{\mathbf{#1}}}}

\newcommand*{\observation}{y} 					

\newcommand*{\gp}{\mathcal{GP}}
\newcommand*{\gpMean}{\mu}

\newcommand*{\parameters}{\mathbf{x}} 
\newcommand*{\ParamSpace}{\set{X}} 

\newcommand*{\state}{s}
\newcommand*{\StateSpace}{\set{S}}
\newcommand*{\action}{a}

\newcommand*{\cost}{c}
\newcommand*{\Cost}{C}
\newcommand*{\trajectory}{\sequence{\state}}
\newcommand*{\stimulus}{v}
\newcommand*{\stimuli}{\sequence{\stimulus}}
\newcommand*{\iReward}{r}       

\newcommand*{\expectation}{\mathbb{E}}

\newcommand*{\normal}{\mathcal{N}}					

\newcommand*{\af}{h} 						

\newcommand*{\iterIdx}{t}

\newcommand*{\primIdx}{i}	

\newcommand*{\nTrajectories}{M}

\newcommand*{\anyfunction}{g}

\let\cite\citep

\title[Adaptive Model Predictive Control by Learning Classifiers]{Adaptive Model Predictive Control by Learning Classifiers}

\author{\Name{Rel Guzman$^2$} \Email{rel.guzmanapaza@sydney.edu.au}\\
 \Name{Rafael Oliveira$^2$} \Email{rafael.oliveira@sydney.edu.au}\\
 \Name{Fabio Ramos$^{1, 2}$} \Email{fabio.ramos@sydney.edu.au}\\
 \addr $^1$NVIDIA, USA, $^2$the University of Sydney, Australia}
 
\begin{document}

\maketitle

\begin{abstract}%
Stochastic model predictive control has been a successful and robust control framework for many robotics tasks where the system dynamics model is slightly inaccurate or in the presence of environment disturbances. Despite the successes, it is still unclear how to best adjust control parameters to the current task in the presence of model parameter uncertainty and heteroscedastic noise. In this paper, we propose an adaptive MPC variant that automatically estimates control and model parameters by leveraging ideas from Bayesian optimisation (BO) and the classical expected improvement acquisition function. We leverage recent results showing that BO can be reformulated via density ratio estimation, which can be efficiently approximated by simply learning a classifier. This is then integrated into a model predictive path integral control framework yielding robust controllers for a variety of challenging robotics tasks. We demonstrate the approach on classical control problems under model uncertainty and robotics manipulation tasks. 
\end{abstract}

\begin{keywords}%
Bayesian methods, Gaussian process, model predictive control, optimisation
\end{keywords}

\section{Introduction}

Reinforcement learning, as a framework, concerns learning how to interact with the environment through experience, while optimal control emphasises sequential decision making and optimisation methods. The boundaries between both fields have been diminished due to deeper understanding and typical applications. Model predictive control (MPC) is an optimisation strategy for behaviour generation that consists of planning actions ahead by minimising costs throughout a horizon. Reinforcement learning and robotics can benefit from MPC by correcting behaviours while constantly estimating hyper-parameters. This controller learning capability can be achieved with data-driven approaches for MPC optimisation \cite{Gorges2017}. 

 We are particularly interested in path integral (PI) control \cite{kappen2005linear}, which is a methodology for solving nonlinear stochastic optimal control problems by sampling trajectories and computing costs. Using such methodology, model predictive path integral control (MPPI) was introduced in \cite{Williams2016}. MPPI enables robots to navigate in stochastic and partially observable environments, for example, in-car racing \cite{Williams2018}. MPPI is a sampling-based and derivative-free method which makes it a simple yet powerful strategy to simulate actions.

Within data-driven approaches, deep reinforcement learning has been successful in solving high-dimensional control problems in simulation~\cite{Duan2016}. The main limitation of deep RL is the need for many interactions with the environment, which can be impractical with a physical system due to costly evaluations~\cite{Peng2017}. An alternative to reduce evaluations is to have a model of the system dynamics, also called a dynamics model or transition model. Modeling an accurate transition model inevitably leads to errors. Even so, by using data-driven approaches, it is possible to reduce the error or adapt to the expected errors in the environment~\cite{Lee2020}.

Data-driven approaches have been proposed for automatic MPC tuning, which can be seen as an intersection between machine learning and control since they make use of the transition model in combination with a learnt model. For example, \cite{sorourifar2021data} presents MPC under uncertainty over model parameters with Bayesian optimisation (BO) that handles constraints of system parameters in a tank reactor. \cite{Lee2020} addresses different environment contexts where a robot's dynamics could change due to a component malfunctioning. Other approaches propose inferring simulation parameters based on data instead of uniform parameter randomisation \cite{Peng2017, Ramos2019}. In another example, the controller optimisation is able to handle heteroscedastic noise for control tasks~\cite{Guzman2020}. Intuitively, heteroscedastic noise is a type of noise that changes with input variables. For example, in stochastic MPC, the noise associated with the stochastic process changes significantly with the temperature hyper-parameter ~\cite{Guzman2020}, making hyper-parameter tuning quite challenging from an optimisation perspective. 

We propose a data-driven approach to optimise a stochastic controller by adapting the transition model parameters to the environment. For costly system evaluations, we use surrogate-based optimisation. Instead of merely optimising objective functions, we optimise an alternative function approximating the more complex, higher-order model to quickly find the local or global optimum. These are used when function evaluations are expensive and noisy due to the environment characteristics. In this context, a surrogate model can amortise the optimisation. Surrogate-based modelling is usually associated with Gaussian processes~\cite{Rasmussen2006}. We consider single-objective optimisation problems with continuous inputs and noisy observations. We work on the case where the noise variance depends on the input. That leads us to heteroscedastic optimisation, which is a more realistic approach than the typical homoscedastic assumption. Heteroscedastic noise can cause problems for the surrogate model and make the optimisation method deviate from the maximum~\cite{Wang2017}. To solve this issue, we aim at finding an optimisation method for stochastic simulations under heteroscedastic noise.

BO uses a Gaussian process \cite{Rasmussen2006} as a surrogate model commonly used for costly black-box functions. BO proposes solutions according to an acquisition function that encodes the trade-off between exploration and exploitation of the objective function. GPs have excellent generalisation properties, but their computational cost can be prohibitive for big data. Additionally, standard GPs provide analytical solutions for posteriors under homoscedastic noise, while heteroscedastic approximations might require computationally expensive approximations.    

An alternative formulation to BO, which allows the utilisation of simple classifiers within the optimisation loop, was proposed in \cite{pmlr-v139-tiao21a} as Bayesian optimisation by Density Ratio Estimation (BORE). The method introduces the concept of relative density ratio, which is used to estimate the expected improvement acquisition function~\citep{Bull2011}. The main advantage of this formulation is that density ratios are bounded between 0 and 1 and can be estimated using any off-the-shelf probabilistic classifier. Classifiers are easy to train and can handle a variety of input noise types, including heteroscedastic, without major modifications to the classification function.   

\textbf{Contributions:} The main contribution of this work is a new robust and adaptive MPC method that automatically estimates distributions of model parameters and MPC hyper-parameters such as the temperature by continuously updating a classifier that acts as a proxy for a Bayesian optimisation step. We demonstrate that the approach provides superior performance in general control problems and manipulation tasks under model uncertainty.

\section{BACKGROUND}
\label{sec:background}

\subsection{Stochastic Model Predictive Control}
\label{subsec:mppi_definition}

Model predictive control resides on the idea of optimising an action sequence up to certain horizon $T$ while minimising the sum of action costs, starting from a current state. MPC returns a next optimal action $a^*$ that is sent to the system actuators. Unlike classical deterministic MPC, stochastic MPC allows disturbances over the states. A stochastic MPC method models disturbances as random variables. At each time step $t$, stochastic MPC generates sequences of perturbed actions $V_t = \{\stimulus_i\}_{i=t}^{t+T}$ where $\stimulus_i = \action_i^* + \epsilon_i$ and $\epsilon_i \sim \normal(0, \sigma_\epsilon^2)$, based on a roll-over sequence of optimal actions $\{\action_i^*\}_{i=t}^{t+T}$ that start at $t=0$. Each action results in a state produced by a transition or dynamics model $s_{t+1} = f\left(s_{t}, a_{t}\right)$, and action sequences result in a state trajectory $\trajectory_t = \{\state_{t+i}\}_{i=1}^T$. Each trajectory has a cumulative cost $\Cost$ determined by instant costs $c$ and a terminal cost $q$:
\begin{align}
\Cost(\trajectory_t)=q\left(s_{t+T}\right) + \sum_{i=1}^{T-1} \cost(s_{t+i})~.
\end{align}
In stochastic MPC, the goal is to minimise the expected $C(S_t)$. The stochastic MPC method known as model predictive path integral (MPPI) and its variations \cite{Williams2016, Williams2018} provide optimal actions for the entire horizon following an information-theoretic approach. Constraints over the states are determined by the transition model, and the actions are constrained according to their limits. After $M$ simulated rollouts, MPPI updates the sequence of optimal actions and weights:
\begin{equation}
\action_{i}^* \leftarrow \action_i^* + \sum_{j=1}^{\nTrajectories} w(\stimuli_t^{j})\epsilon_i^{j}\,, \qquad
w(\stimuli_t) = \frac{1}{\eta} \exp\left(-\frac{1}{\lambda}\left(\Cost(\trajectory_t) + \frac{\lambda}{\sigma_\epsilon^2} \sum_{i=t}^{t+T}\action_{i}^*\cdot\stimulus_{i}\right)\right)\,,
\label{eq:mppi_eq}
\end{equation}
where $j \in \{1,\ldots, \nTrajectories\}$ and $\eta$ is a normalisation constant so that $\sum_{j=1}^\nTrajectories w(\stimuli_t^j) = 1$. The hyper-parameter temperature $\lambda \in \R^+$, $\lambda \to 0$ leads to more peaked distributions for actions~\cite{Williams2018a}. The hyper-parameter $\sigma_\epsilon^2$ is the control variance, which leads to more varying and forceful actions when it is increased.

\subsection{Classic Bayesian Optimisation}

Bayesian optimisation (BO) has been widely applied in robotics and control to optimise black-box (i.e., derivative-free) functions that are costly to evaluate in applications such as robotics. That is due to the usual cost of running experiments in real robots. We use BO to perform global optimisation in a given search space $\ParamSpace$. BO uses a Gaussian process (GP) \cite{Rasmussen2006} as a surrogate model $\mathcal{M}$ to internally approximate an objective function $\anyfunction:\ParamSpace \to \mathbb{R}$. BO defines an optimisation problem $\mathbf x^* \in \argmax_{\mathbf x \in \ParamSpace} \anyfunction(\mathbf x)$. Then, given a set of collected observations $\mathcal{D}_{1:t}$, BO constructs a surrogate model that provides a posterior distribution over the objective function $g$. This posterior is used to construct the acquisition function $h$, which measures both performance and uncertainty of unexplored points. The next step is optimising the acquisition function, obtaining a sample $(\mathbf{x}_t, y_t)$. BO is summarised in \autoref{alg:bo}.

A popular acquisition function in the BO literature is the expected improvement (EI) \citep{Bull2011}. At iteration $\iterIdx$, one can define $\observation_\iterIdx^* := \max_{\primIdx<\iterIdx} \observation_\iterIdx$ as an optimal incumbent. The expected improvement is then defined as:
\begin{equation}
\af_{\operatorname{EI}}(\parameters|\dataset_{\iterIdx-1}) := \expectation[\max\left\{0,f(\parameters)-\observation_{\iterIdx}^*\right\}|\dataset_{\iterIdx-1}]~.
\label{eq:ei}
\end{equation}
In the case of a GP prior on $f|\dataset_{\iterIdx-1}\sim\gp(\gpMean_{\iterIdx-1}, k_{\iterIdx-1})$, for any point $\parameters$ where the predictive standard deviation of the GP is non-zero, i.e., $\sigma_{\iterIdx-1}(\parameters) = \sqrt{k_{\iterIdx-1}(\parameters, \parameters)} > 0$, the EI is given by:
\begin{equation}
\af_{\operatorname{EI}}(\parameters|\dataset_{\iterIdx-1}) =
(\gpMean_{\iterIdx-1}(\parameters)-\observation_{\iterIdx}^*)\Psi(s_\iterIdx) + \sigma_{\iterIdx-1}(\parameters)\psi(s_\iterIdx)\,,
\end{equation}
where $s_\iterIdx := \frac{\gpMean_{\iterIdx-1}(\parameters)-\observation_{\iterIdx}^*}{\sigma_{\iterIdx-1}(\parameters)}$, if $\sigma^2_{\iterIdx-1}(\parameters) := k_{\iterIdx-1}(\parameters,\parameters) > 0$. For points where $\sigma_{\iterIdx-1}(\parameters) = 0$, i.e., there is no posterior uncertainty, we simply have $\af_{\operatorname{EI}}(\parameters|\dataset_{\iterIdx-1}) = 0$. Here $\Psi(s_\iterIdx)$ and $\psi(s_\iterIdx)$ denote the cumulative and probability density functions of the standard normal distribution evaluated at $s_\iterIdx$.

\subsection{Bayesian Optimisation by Learning Classifiers}

BO is hindered by the GP surrogate model, for which it has limitations such as cubic computational cost in training and not being directly amenable to handle variable noise structures such as heteroscedasticity. The extensions proposed to address those issues are restrained by the necessity to ensure analytical tractability and typically make strong and oversimplifying assumptions. For example, \cite{roy2013variational} proposed an heteroscedastic BO approach that uses a variational approximation that can be expensive to compute. BO by Density Ratio Estimation (BORE) \cite{pmlr-v139-tiao21a} was proposed based on a reformulation of the expected improvement acquisition function (\autoref{eq:ei}) and bypassed the challenges of analytical tractability in GP-based approaches. BORE works by selecting points according to a density ratio similar to the Tree-structured Parzen Estimator (TPE) proposed by~\citet{Bergstra2011}. TPE divides the observations, based on some quantile hyper-parameter, into a first group that gave the best scores and a second group containing the rest. Then, the goal is to find inputs that are more likely to be in the first group. In order to propose a new sampling point, TPE computes the ordinary density ratio from \autoref{eq:ratio} between the probability $a(\mathbf{x})$ of being in the first group, and the probability $b(\mathbf{x})$ of being in the second group. Instead, BORE uses the $\gamma$-relative density ratio $r_\gamma$. The ordinary and $\gamma$-relative density ratios are specified by:

\begin{minipage}{0.4\textwidth}
\begin{equation}
r_0(\mathbf{x})=a(\mathbf{x}) / b(\mathbf{x})\,, \label{eq:ratio}
\end{equation} 
\end{minipage}%
\begin{minipage}{0.5\textwidth}
\begin{equation}
r_{\gamma}(\mathbf{x}):=\frac{a(\mathbf{x})}{\gamma a(\mathbf{x})+(1-\gamma) b(\mathbf{x})}\,. \label{eq:relativeratio}
\end{equation}
\end{minipage}\\

BORE approximates the $\gamma$-relative density ratio to a binary class posterior probability as $r_{\gamma}(\mathbf{x}) \simeq \gamma^{-1} \pi(\mathbf{x})$, where $\pi$ computes the probability of $\mathbf{x}$ belonging to a positive class $\pi(\mathbf{x})=p(z=1 \mid \mathbf{x})$. The binary label $z$ introduced here denotes a negative or positive class, and its meaning corresponds to whether the point should be selected or not. In BORE, given a maximisation objective, we set $z := \mathbb{I}[\observation \geq \tau]$, indicating of whether the corresponding observation $\observation$ at a point $\parameters$ is above the $\gamma$th quantile $\tau$ of the (empirical) observations distribution, i.e., $\gamma = p(\observation \geq \tau)$. In the end, computing the acquisition function $h$ from the classical BO method (\autoref{alg:bo}) is reduced to classifier training. 

\begin{figure*}[t]
\begin{minipage}{0.47\linewidth}
\begin{algorithm2e}[H]
\caption{Bayesian Optimisation}
\label{alg:bo}
\SetAlgoLined
\SetKwInOut{Input}{input}\SetKwInOut{Output}{output}
\footnotesize
\Input{ Sampling iterations $n$; search space $\ParamSpace$; hyper-parameters of $\af$}
\Output{ $\left(\mathbf{x}^{*}, y^{*}\right)$ }
\For{$t = 1$ \KwTo $n$}{
    Fit a GP model $\mathcal{M}$ on the data $\mathcal{D}_{t-1}$\\
    Find $\mathbf{x}_{t}=\argmax_{\mathbf x \in \ParamSpace} \af(\mathbf{x},  \mathcal{M}, \mathcal{D}_{t-1})$\\
    $y_t \leftarrow f(\mathbf{x}_t)$\\
	$\mathcal{D}_{t} \leftarrow \mathcal{D}_{t-1} \cup \{(\mathbf{x}_t, y_t)\}$
 }
\end{algorithm2e}
\end{minipage}%
\hfill%
\begin{minipage}{0.49\linewidth}
\begin{algorithm2e}[H]
\caption{BORE}
\label{alg:bore}
\SetAlgoLined
\SetKwInOut{Input}{input}\SetKwInOut{Output}{output}
\footnotesize
\Input{Sampling iterations $n$; search space $\ParamSpace$; $\gamma \in (0,1)$; classifier to train $\pi : \ParamSpace \to [0,1]$}
\Output{ $\left(\mathbf{x}^{*}, y^{*}\right)$ }
\For{$t = 1$ \KwTo $n$}{
  Train the probabilistic classifier $\pi$\\
  Find $\mathbf{x}_{t} = \arg \max _{\mathbf{x} \in \ParamSpace} \pi(\mathbf{x})$\\
  $y_t \leftarrow f(\mathbf{x}_t)$\\
  $\mathcal{D}_{t} \leftarrow \mathcal{D}_{t-1} \cup \{(\mathbf{x}_t, y_t)\}$
}
\end{algorithm2e}
\end{minipage}
\end{figure*}

As shown in \autoref{alg:bore}, the optimisation depends on the hyper-parameter $\gamma \in (0, 1)$, which influences the exploration-exploitation trade-off. A smaller $\gamma$ encourages exploitation. In this work, we anneal $\gamma$ until it reduces to a minimum value close to 0. The reason is to induce more exploration initially and avoid local minima as much as possible.
\section{Methodology}
\label{sec:method}

\subsection{Dynamics Model Uncertainty}

The dynamics of the environment is modelled as a Markovian transition model. We consider a transition model with states $\mathbf \state \in \StateSpace$ and admissible actions $\mathbf a \in \mathcal{A}$. The state follows Markovian dynamics, $\mathbf s_{t+1}=f\left(\mathbf s_{t}, \mathbf a_{t}\right)$, with a transition function $f$ and a reward function $\iReward$ that evaluates the system performance given a state and action $\iReward : \mathcal{S} \times \mathcal{A} \to \R$. That transition model can be learned or assumed from expert knowledge. 

We propose adapting stochastic MPC to the environment by exposing the robot to different possible scenarios by defining a transition model parameterised by a random variable $\theta$. To find an optimum $\theta$, we add randomisation at each MPC trajectory rollout. We define a random vector of transition model parameters $\boldsymbol{\theta}$ and adapt them to the stochastic MPC controller. Each transition model parameter follows a probability distribution parameterised by $\boldsymbol{\psi}$: 
\begin{equation}
\boldsymbol{\theta} \sim p_{\boldsymbol{\theta}}(\boldsymbol{\theta}; \boldsymbol{\psi}), \qquad \mathbf{s}_{t+1} = f\left(\mathbf{s}_{t}, \mathbf{a}_t + \boldsymbol{\epsilon}_t, \boldsymbol{\theta} \right)~,
\label{eq:modelparameter}
\end{equation}
where $\mathbf{s}_{t}$ is the state obtained at time $t$, and $\mathbf{a}_t + \boldsymbol{\epsilon}_t$ is the perturbed action as described in stochastic MPC~(\autoref{subsec:mppi_definition}). Note that $\boldsymbol{\theta}$ is now an input to the dynamics model. Finally, optimal actions found by MPC are sent to the system using the dynamics model $f$.

\subsection{An Adaptive Control Formulation}

We do not directly aim at finding parameters that match the observed dynamics as we would do in system identification \cite{Romeres2016}. Instead, we adapt model parameter distributions to the controller, which means those resulting distributions may be close to their true values. The uncertainty of those parameters would allow the controller to adapt under different environment circumstances or characteristics, such as the size of an obstacle or the length of a robot component.

\begin{figure*}[t]
    \centering
    \includegraphics[width=0.7\columnwidth]{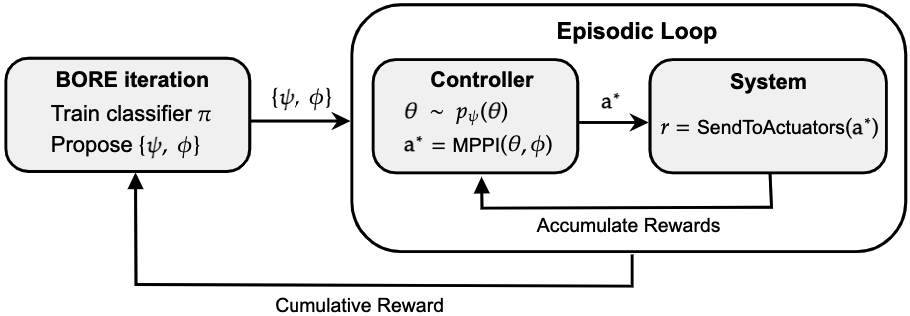}
    \caption{An overview of adaptive MPC by learning classifiers.}
    \label{fig:overview}
\end{figure*}

We use the optimal MPC hyper-parameters by adapting them to the transition model parameters. In the case of MPPI \cite{Williams2018a}, the hyper-parameters considered are $\lambda$ and $\sigma_\epsilon$ as described in \autoref{subsec:mppi_definition}. All controller hyper-parameters are collectively described as $\boldsymbol{\phi}$. We introduce the reinforcement learning objective of optimising the episodic cumulative reward $R$. An overview of the proposed framework is shown in \autoref{fig:overview}. The objective is to solve the reward optimisation problem where we jointly estimate $\boldsymbol{\psi}$, which are hyper-parameters for the dynamics model parameter distribution $p_{\boldsymbol{\psi}}(\boldsymbol\theta)$, and the controller hyper-parameters $\boldsymbol{\phi}$:
\begin{equation}
\boldsymbol{\psi}^{\star}, \boldsymbol{\phi}^{\star}=\argmax_{\{\boldsymbol{\psi}, \boldsymbol{\phi}\}}\, R(\boldsymbol{\psi}, \boldsymbol{\phi})~.
\label{eq:optproblem}
\end{equation}

To perform the optimisation in \autoref{eq:optproblem}, we adopt a surrogate-based optimisation method that can handle noisy gradient-free functions that are common in control tasks. To this end, we use BORE since it provides a number of advantages over traditional GP-based BO, as discussed previously, and it can handle heteroscedasticity without modifications to the core method. Heteroscedastic noise is common when tuning stochastic MPC in control problems \cite{Guzman2020}. A difference between GP-based BO and BORE is that BO filters noise directly via the GP model, while BORE relies on the classifier to account for label noise. Therefore, to mitigate noise effects, we propose to optimise the objective function averaged over a small number $n_e$ of episodes.

To better understand the optimisation framework, \autoref{alg:proposal} describes how to estimate the optimal controller and dynamics hyper-parameters $\mathbf x^* = \{\boldsymbol{\psi}^{*}, \boldsymbol{\phi}^{*}\}$ using a binary classifier. Following the reinforcement learning literature, we define the cumulative reward $R = \sum_{i=1}^{n_s} r_i$, where $n_s$ is the number of time steps in an episode, and set our goal as maximising the expected cumulative reward $g = \mathbb{E}[R]$. We compute an empirical expected cumulative reward $g$ by averaging $R$ over $n_e$ episodes. The classifier $\pi_t$ is trained by first assigning labels $\{z_k\}_{k=1}^{t}$ to the data observed up to the current iteration $t$. For training, the classifier uses the auxiliary dataset:
\begin{equation}
\left\{\left(\boldsymbol{\psi}_k, \boldsymbol{\phi}_k, z_{k}\right)\right\}_{k=1}^{t}~,
\label{eq:auxiliary_dataset}
\end{equation}
where the labels are obtained by separating the observed data according to $\gamma \in (0, 1)$ by computing the $\gamma$th quantile of $\left\{g_{k}\right\}_{k=1}^{t}$:
\begin{equation}
\tau \leftarrow \Phi^{-1}(\gamma), \qquad
z_{k} \leftarrow \mathbb{I}[g_{k} \geq \tau] \quad \text{ for } k=1, \ldots, t~.
\label{eq:separation}
\end{equation}
The exploration-exploitation trade-off is balanced by $\gamma$, with small $\gamma$ encouraging more exploitation. Instead of keeping $\gamma$ fixed, we define a strategy that first explores the search space with an initial high $\gamma_1$ that decays linearly across the iterations until a final $\gamma_n$. Inputs predicted as positive labels are considered to have a higher reward, and one of them is selected by maximising the classifier output:
\begin{equation}
\boldsymbol{\psi}_t, \boldsymbol{\phi}_t = \argmax_{\{\boldsymbol{\psi}, \boldsymbol{\phi}\} \in \ParamSpace} \pi_{t-1}(\boldsymbol{\psi}, \boldsymbol{\phi})~.
\label{eq:max_acquisition}
\end{equation}
Note that this is equivalent to acquisition function maximisation in conventional BO and allows the method to suggest candidate solutions efficiently. For better performance, the maximisation can be carried out with a global optimisation method \citep[e.g.,][]{Arnold2010}. 

\begin{figure*}[h]
\begin{minipage}{1.0\linewidth}
\begin{algorithm2e}[H]
\caption{Adaptive MPC by learning classifiers}
\label{alg:proposal}
\SetAlgoLined
\SetKwInOut{Input}{input}\SetKwInOut{Output}{output}
\footnotesize
\Input{Sampling iterations $n$ \\
Search space $\ParamSpace$\\
Initial $\gamma_1$ and final $\gamma_n$\\
Probabilistic binary classifier $\pi : \ParamSpace \to [0,1]$}
\Output{ $\left(\boldsymbol{\psi}^{*}, \boldsymbol{\phi}^{*}, g^{*}\right)$ }
\For{$t = 1$ \KwTo $n$}{
  $\gamma_t = \gamma_1 - \frac{t-1}{n-1}(\gamma_1 - \gamma_n)$ \hspace{35mm} \textcolor{gray}{// linear $\gamma$ decay}\\
  $\tau \leftarrow \Phi^{-1}(\gamma_t)$ \hspace{50mm} \textcolor{gray}{// compute $\gamma_t$-th quantile of $\left\{g_{k}\right\}_{k=1}^{t}$}\\
  $z_{k} \leftarrow \mathbb{I}\left[g_{k} \geq \tau\right] \text { for } k=1, \ldots, t-1$ \hspace{19.5mm} \textcolor{gray}{// assign labels to the observed data points}\\
  Train the classifier $\pi_{t-1}$ using $\left\{\left(\boldsymbol{\psi}_k, \boldsymbol{\phi}_k, z_{k}\right)\right\}_{k=1}^{t-1}$ \hspace{4.5mm} \textcolor{gray}{// acquisition function according to BORE}\\
  $\boldsymbol{\psi}_t, \boldsymbol{\phi}_t = \arg \max _{\{\boldsymbol{\psi}, \boldsymbol{\phi}\} \in \ParamSpace} \pi_{t-1}(\boldsymbol{\psi}, \boldsymbol{\phi})$ \hspace{17mm} \textcolor{gray}{// estimate new hyper-parameters and model parameters}\\
  \For{$j = 1$ \KwTo $n_e$}{
        $R_j^{(t)} = 0$\\
        \For{$i = 1$ \KwTo $n_s$}{
            $\boldsymbol{\theta} \sim p_{\boldsymbol{\theta}}(\boldsymbol{\theta}; \boldsymbol{\psi_t})$ \hspace{23.5mm} \textcolor{gray}{// sample from the parameter distributions} \\
            $a_{i}^* = \text{MPC}(f, \boldsymbol{\theta}_t, \boldsymbol{\phi}_t)$  \hspace{15mm} \textcolor{gray}{// use estimated parameter distributions in the new transition model} \\
            $r_{i} = \text{SendToActuators}(a_{i}^*)$ \hspace{8.5mm} \textcolor{gray}{// evaluate the first action to take in the optimal trajectory} \\
            $R_j^{(t)} \mathrel{+}= r_{i}$ \hspace{28mm} \textcolor{gray}{// accumulate rewards}
        }
    }
  Decrease $\gamma$ to reduce explorability \\
  $g_t = 1/n_e \sum_{j=1}^{n_e}\left[R_j^{(t)}\right]$\\
  $\mathcal{D}_{t} \leftarrow \mathcal{D}_{t-1} \cup \{(\boldsymbol{\psi}_t, \boldsymbol{\phi}_t, g_t)\}$
}
\end{algorithm2e}
\end{minipage}
\end{figure*}

\subsection{Choosing the Classifier}

The probabilistic binary classifier in \autoref{eq:max_acquisition} has to be chosen considering the observation noise in the task. Some methods used in \citet{pmlr-v139-tiao21a} are XGBoost, multi-layer perceptron (MLP), and Random Forests (RF). For example, as an ensemble method, RF combines decision trees via bagging. The number of decision trees should be sufficiently large to reduce classification variance without increasing the bias. We highlight the study of this method since it is an out-of-the-box classification method.

\section{Experiments}
\label{sec:experiments}

We consider the problem of optimising the function $R(\mathbf{x})=g\left(\mathbf{x}\right) + \epsilon$ with heteroscedastic input-dependent noise $\epsilon \sim \mathcal{N}(0,\sigma_\epsilon^2(\mathbf{x}))$, where $\sigma_\epsilon^2(\mathbf{x})$ denotes an input-dependent noise variance. In these experiments the variables we optimise are the controller hyper-parameters $\boldsymbol{\phi}$, and the variable $\boldsymbol{\psi}$ that parameterises  transition model parameter distributions. We aim at optimising the true noise-free function $g$ although we only have access to the cumulative reward $R$. We use $R$ as the objective to optimise in the simulator experiments. We evaluate the performance of the proposed adaptive model predictive control framework in several experiments described below. 

\subsection{Simulation Experiments}

In this section, we assess the performance of the adaptive controller framework in control and robotic tasks. Specifically, we ran tests in simulated environments, such as Pendulum, Half-Cheetah, and Fetchreach from OpenAI\footnote{OpenAI Gym: \url{https://gym.openai.com}} with dense reward functions. The same functions taken from \citet{Guzman2020} are used to compute instant costs for MPPI. We also experimented with the reaching task for the Panda robot environment from \citet{bhardwaj2021storm}, with a single obstacle, a fixed target location, and a fixed initial robot position. MPPI trajectory evaluations are done on the GPU, which helped overcome efficiency issues.

\begin{wrapfigure}{r}{0.3\textwidth} 
    \centering
    \vspace{-.4cm}
    \includegraphics[width=0.3\columnwidth]{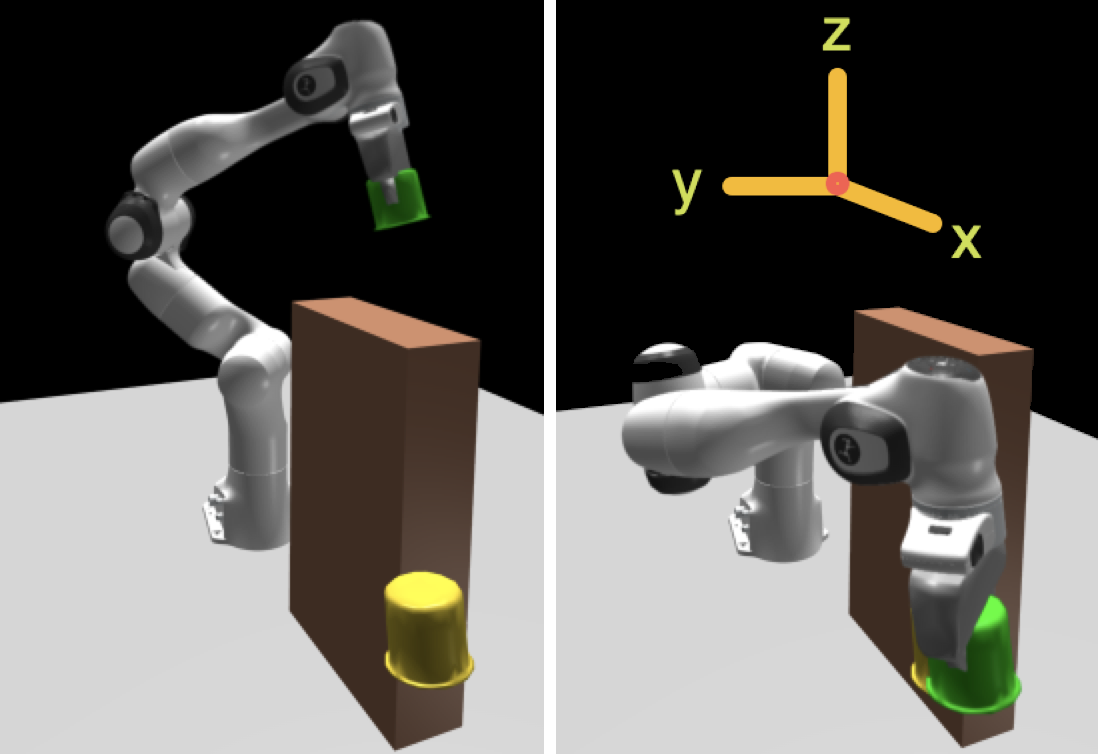}
    \caption{Panda robot setup}
    \label{fig:robot}
    \vspace{-.2cm}
\end{wrapfigure}
The purpose of the Panda task shown in \autoref{fig:robot} is to reach the yellow target while avoiding obstacle collision. The obstacle has true length $x=0.3$, width $y=0.1$, and height $z=0.6$. We assume partial observability for such obstacle dimension sizes and attempt to infer them as part of the transition model parameters for which we define search spaces shown in \autoref{tab:pranges}. The transition model parameter $l$ is the rod length for Pendulum, $\kappa_m$ is the mass scaling factor for all the links in Half-Cheetah, and $\kappa_d$ is a damping ratio scaling factor for all components in Fetchreach. For Panda, we optimise the obstacle dimensions $x$, $y$, and $z$. Each transition model parameter is a random variable parameterised by $\vec\psi$, for example $\vec\psi = \{\kappa_{d,\mu}, \kappa_{d,\sigma}\}$ are damping ratio mean and damping ratio standard deviation for the Fetchreach.

\begin{center}
\begin{table}[t]
\small
    \centering
    \setlength{\tabcolsep}{0.3em} 
    \renewcommand{\arraystretch}{1.1}
    \begin{tabular}{|c|c|c|c|c|c|c|c|}
        \hline
        \rowcolor{gainsboro}
        Environment   & $n_e$ & $T$ & $M$   & Control hyp. & \multicolumn{2}{c|}{Distribution parameter range} & True parameter \\
        \hline
        Pendulum     & 1 & 10 & 10 & \shortstack{$\lambda \in {[}0.01, 50{]}$ \\ $\sigma_\epsilon \in {[}1.0, 10{]}$}  & \shortstack{$\mu_l \in {[}0.5, 1.6{]}$\vspace{1mm}} & \shortstack{$\sigma_l \in  {[}0.001, 0.1{]}$\vspace{1mm}} & \shortstack{$l=1.0$ \vspace{1.6mm}} \\
        \hline
        Half-Cheetah & 18 & 15 & 10 & \shortstack{$\lambda \in {[}0.01, 1.0{]}$ \\ $\sigma_\epsilon \in {[}0.05, 2.0{]}$}  & \shortstack{$\kappa_{m,\mu} \in {[}0.2, 2.0{]}$\vspace{1mm}}   & \shortstack{$\kappa_{m,\sigma} \in  {[}0.001, 0.1{]}$\vspace{1mm}} & \shortstack{$\kappa_{m}=1.0$ \vspace{1.6mm}}     \\
        \hline
        Fetchreach & 90 & 12 & 3 & \shortstack{$\lambda \in {[}0.01, 0.03{]}$ \\ $\sigma_\epsilon \in {[}0.001, 0.5{]}$}  & \shortstack{$\kappa_{d,\mu} \in {[}1.0, 50{]}$ \vspace{1mm}}   & \shortstack{$\kappa_{d,\sigma} \in  {[}0.001, 0.6{]}$ \vspace{1mm}}   & \shortstack{$\kappa_{d}=1.0$ \vspace{1.6mm}}  \\
        \hline
        \shortstack{Panda\vspace{4mm}} & \shortstack{10\vspace{4mm}} & \shortstack{150\vspace{4mm}} & \shortstack{20\vspace{4mm}} & \shortstack{$\lambda \in {[}0.01, 2.0{]}$ \vspace{4mm}}  & \shortstack{$x_{\mu} \in {[}0.3, 0.32{]}$\\ $y_{\mu} \in {[}0.1, 0.12{]}$ \\ $z_{\mu} \in {[}0.6, 0.62{]}$} &
          \shortstack{ $x_{\sigma} \in {[}0.001, 0.05{]}$\\ $y_{\sigma} \in {[}0.001, 0.01{]}$ \\ $z_{\sigma} \in {[}0.001, 0.03{]}$} & \shortstack{$x=0.3$\\ $y=0.1$ \\ $z=0.6$ \vspace{1mm}} \\
        \hline
    \end{tabular}
    \caption{Ranges of hyper-parameters and model parameters.}
    \label{tab:pranges}
\end{table}
\end{center}

\vspace{-1.5cm}
\subsection{Method Configuration}

We compare the proposed adaptive MPC framework with other surrogate-based methods used in robotics. All the compared methods are configured as follows.\\

\begin{wrapfigure}{r}{0.3\textwidth} 
\vspace{-.4cm}
\includegraphics[width=0.28\columnwidth]{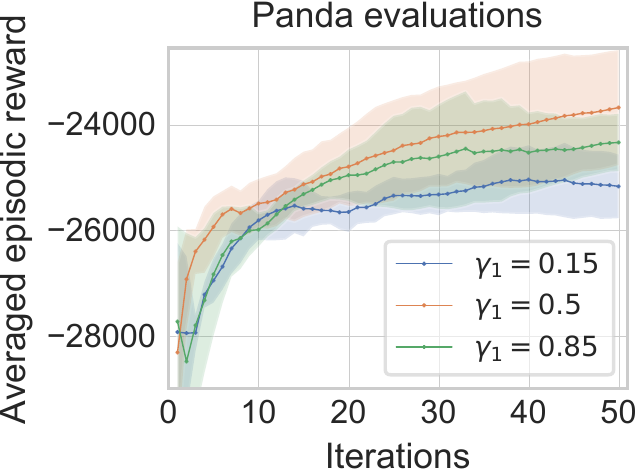}
\caption{Some $\gamma$ evaluations} \label{fig:gammas}
\vspace{-.2cm}
\end{wrapfigure}
\noindent\textbf{Adaptive MPC framework:} First, we have to determine a classifier that can deal with the stochasticity of robotic tasks. We compare two probabilistic classifiers: Random Forest (RF) with 50 decision trees denoted as BORE-RF, and Multi-layer Perceptron (MLP) classifier denoted as BORE-MLP with 2 hidden layers, each with 32 units, ReLU activations and sigmoid for the output layer. The weights were optimised for 1000 epochs using binary cross-entropy loss and ADAM optimiser with a batch size of 32. Second, we start by exploring with a $\gamma_1=0.5$ that decays linearly across the iterations until a reasonable final $\gamma_n=0.05$. We compare different initial $\gamma_1$ values in \autoref{fig:gammas} for the Panda task. With a low $\gamma_1$, BORE could stay stuck in some local minimum, and with a higher $\gamma_1$, BORE would do more exploration first before exploiting some region. A reasonable initial value is $\gamma_1=0.5$, corresponding to the median, which has shown an optimal compromise according to the preliminary results in \autoref{fig:gammas}. Finally, we set a parameter distribution $p_{\boldsymbol{\theta}}$ with positive support since we deal with physics variables (mass, damping ratio) and sizes. We choose the gamma distribution $\Gamma(\alpha, \beta)$, and we transform the provided mean $\mu$ and standard deviation $\sigma$ via $\alpha=\frac{\mu^2}{\sigma^2}$ and $\beta=\frac{\mu}{\sigma^2}$. \\

\noindent \textbf{BO methods:} We compare the proposed method with the traditional homoscedastic B0$_{\operatorname{homo}}$ and a heteroscedastic BO$_{\operatorname{hetero}}$ version from \citet{Guzman2020}. We collected 400 data points for the control and robotic tasks over the search spaces shown in \autoref{tab:pranges}. Then, using such data, we optimise BO's hyper-parameters via maximum GP marginal likelihood optimisation. That marginal likelihood and the acquisition function are optimised using multi-start L-BFGS-B. We used a UCB acquisition function $\af_{\operatorname{UCB}}(\vec{x})=\mu(\vec{x}) + \delta \sigma(\vec{x})$ with balance factor $\delta=3.0$. For both BO, we used the anisotropic squared exponential kernel $k(\mathbf x, \mathbf x') =  \sigma_n\exp\left(-1/2(\mathbf x - \mathbf x')^T \operatorname{diag}(\boldsymbol{\ell})^2(\mathbf x - \mathbf x')\right)$.\\

\noindent \textbf{Other methods:} We use TPE optimisation \citep{Bergstra2011} with quantile value $0.5$ since it is what BORE is based on, and finally, we use covariance matrix adaptation evolution strategy (CMA-ES) \cite{Arnold2010} as a non-BO baseline set with $\sigma_0=10$ and population size $2$. CMA-ES has been widely used for hyper-parameter tuning in robotics \citep{modugno2016learning, sharifzadeh2021maneuverable}.\\

\begin{figure}[t]
\centering
\includegraphics[width=0.9\textwidth]{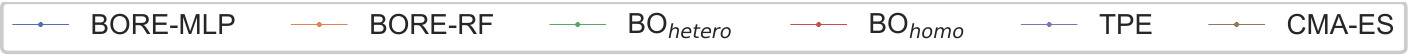}\\
\vspace{2mm}
\includegraphics[height=0.225\textwidth]{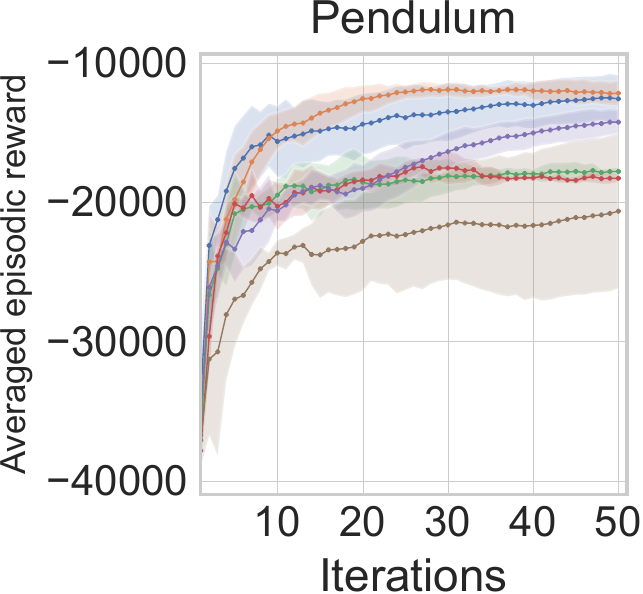}
\hspace{1mm}
\includegraphics[height=0.225\textwidth]{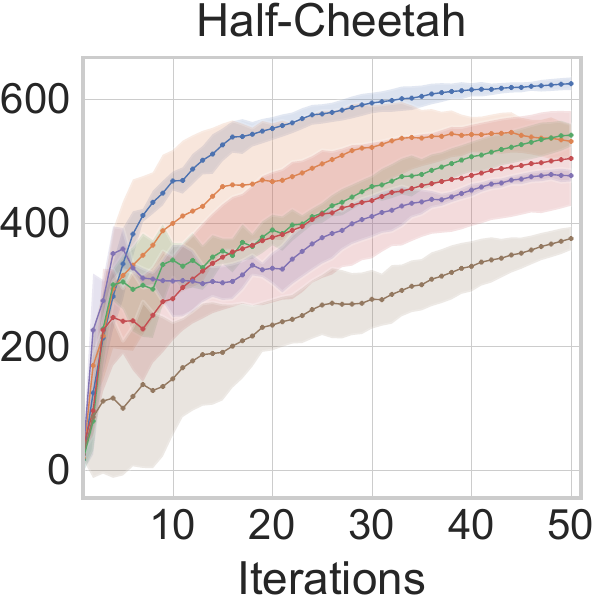}
\hspace{1mm}
\includegraphics[height=0.225\textwidth]{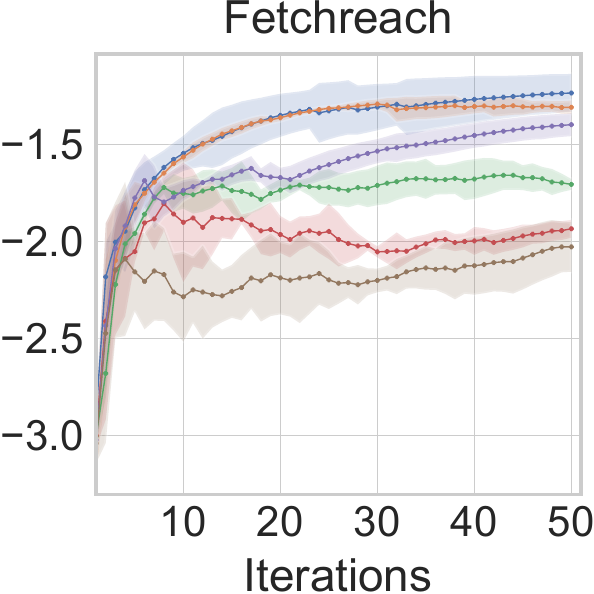}
\hspace{1mm}
\includegraphics[height=0.225\textwidth]{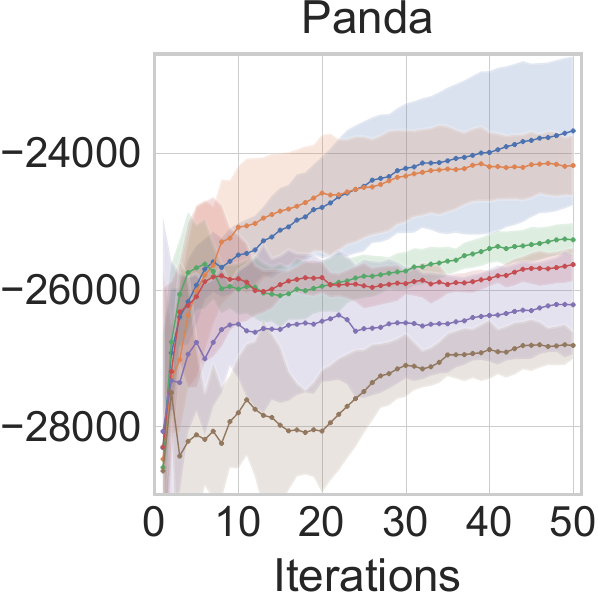}
\caption{Averaged cumulative rewards per iteration where the shaded areas correspond to 1.5 standard deviations. Each method started at a point with minimum expected cumulative reward.}
\label{fig:rewardcurves}
\end{figure}

\subsection{Optimisation Assessment}

To quantitatively assess optimisation performance, we report averaged cumulative rewards, defined as $\frac{1}{t'} \sum_{i=1}^{t'} g_i$ from $t'=1$ to $t'=n$. To account for the stochasticity of evaluations, we repeat those 50 iterations 3 times. The result of repeating gives averaged cumulative rewards with their respective standard deviations. We use the cumulative rewards $R$ to approximate $g$ by averaging over $n_e=10$ episodes. Each episode consists of 480 time steps for Panda and 200 time steps for the other tasks. We compare the averaged cumulative reward against the number of iterations. \autoref{fig:rewardcurves} shows that BO$_{\operatorname{homo}}$ and BO$_{\operatorname{hetero}}$ perform similarly mainly in Pendulum because of homoscedastic noise across the search space. However, BO$_{\operatorname{homo}}$ tends to converge to a local minimum in other tasks, which is expected since it does not account for heteroscedasticity. It is possible to achieve better or equal results with TPE, although it seems to also get stuck since it only divides observations based on the output and chooses the best next point without considering unseen regions. Both BO versions are being outperformed by BORE-MLP and BORE-RF. BORE-MLP converges quite faster to an optimum in most tasks. In the PANDA environment, the difference is higher, and it suggests that the proposed framework performs better in higher-dimensional problems.

\begin{table}[t]
\centering
\small
\begin{tabular}{| c >{\columncolor{yellow!18}} c >{\columncolor{yellow!18}} c >{\columncolor{gray!18}} c >{\columncolor{orange!18}} c >{\columncolor{orange!18}} c >{\columncolor{green!18}} c >{\columncolor{green!18}} c >{\columncolor{cyan!18}} c >{\columncolor{cyan!18}}c |}
\toprule
       Method & $R_{max}$ & $R_{\sigma}$ & $\lambda$ & $x_{\mu}$ & $x_{\sigma}$ & $y_{\mu}$ & $y_{\sigma}$ & $z_{\mu}$ & $z_{\sigma}$ \\
\midrule
BORE-MLP & -21106.36 & 724.10 & 1.65 & 0.3104 & 0.0010 & 0.1000 & 0.0010 & 0.6000 & 0.0058 \\
\hline
BORE-RF & -21891.82 & 291.64 & 1.67 & 0.3036 & 0.0431 & 0.1001 & 0.0013 & 0.6145 & 0.0234 \\
\hline
BO$_{\operatorname{hetero}}$ & -23025.47 & 162.22 & 1.73 & 0.3185 & 0.0500 & 0.1200 & 0.0028 & 0.6145 & 0.0118 \\
\hline
BO$_{\operatorname{homo}}$ & -22870.14 & 158.46 & 2.00 & 0.3200 & 0.0010 & 0.1000 & 0.0010 & 0.6200 & 0.0010 \\
\hline
TPE & -23438.34 & 121.26 & 1.63 & 0.3124 & 0.0159 & 0.1152 & 0.0045 & 0.6047 & 0.0067 \\
\hline
CMA-ES & -23779.35 & 481.55 & 1.47 & 0.3200 & 0.0010 & 0.1200 & 0.0010 & 0.6200 & 0.0142 \\
\bottomrule
\end{tabular}
    \caption{Maximum reward found at the last iteration for the Panda task.}
    \label{tab:franka_results}
\end{table}

\subsection{Evaluating the Optima}

The previous section emphasised the proposed MPC framework and its ability to explore efficiently compared to other optimisation methods. This section describes the optima found by the methods in the Panda environment, where the improvement is more noticeable.  In \autoref{tab:franka_results}, we show the control hyper-parameters $\vec\phi=\{\lambda\}$ and the transition model parameters $\vec\psi = \{x_{\mu}, x_{\sigma}, y_{\mu}, y_{\sigma}, z_{\mu}, z_{\sigma}\}$ that give the maximum reward $R_{\operatorname{max}}$ at the last iteration after running each method for 50 iterations. $R_{\sigma}$ is the observed standard deviation of the reward at the respective iteration. BORE-MLP is able to find an optimum close to the one found by BORE-RF. $R_{\sigma}$ is higher for BORE-MLP as the method is still exploring new unseen regions at the end, and it can still improve its current maximum. The table also shows the optimised parameters for the distribution-based sizes: orange for the length $x$, green for the width $y$, and cyan for the height $z$. BORE-MLP and almost all the other methods found that considering more uncertainty in the obstacle height $z$ would provide a higher reward, which is understandable considering that the gripper could find convenient trajectories by moving over the obstacle. The most relevant dimension size is the width $y$, since a wrong $y$ would result in obstacle collision. Meanwhile, all methods can allow more uncertainty about the length of the obstacle as it does not affect the collision. Most methods converge to a similar controller hyper-parameter $\lambda$.

\section{Conclusion}
\label{sec:conclusion}
This paper presented an adaptive variant of model predictive control that automatically estimates model parameter distributions and tunes MPC hyper-parameters within a Bayesian optimisation framework. In contrast to previous approaches, our formulation is the first to show that global optimisation can be accomplished by learning a classifier that estimates density ratios. 
We studied the empirical performance of the framework with different classifiers and against benchmark BO versions. The proposed method was able to surpass the performance of the traditional BO and a heteroscedastic BO variation. Our results indicate the flexibility of using density-ratio estimation to optimise MPC and how it can impact the performance of MPC in control and robotic tasks under dynamics model uncertainty. Future research directions include obtaining theoretical results on the effects of heteroscedasticity, and we could explore alternative non-normal distributions for the actions that could be more suitable for the tasks.

\bibliography{rl_bo_references}

\end{document}